\documentclass[runningheads]{llncs}
\usepackage{graphicx}

\usepackage{tikz}
\usepackage{comment}
\usepackage{graphicx}
\usepackage{amsmath}
\usepackage{amssymb}
\usepackage{array,booktabs}
\usepackage{bm}
\usepackage{color}
\usepackage{multirow}
\usepackage{bbding}
\definecolor{Emerald}{rgb}{0.31, 0.78, 0.47}
\definecolor{Denim}{rgb}{0.08, 0.38, 0.74}
\definecolor{Green(ryb)}{rgb}{0.4, 0.69, 0.2}
\definecolor{Iris}{rgb}{0.35, 0.31, 0.81}
\usepackage[accsupp]{axessibility}  
\usepackage[pagebackref,breaklinks,colorlinks]{hyperref}
\usepackage{diagbox}
\usepackage{appendix}

\begin{document}
\pagestyle{headings}
\mainmatter
\def\ECCVSubNumber{4059}  

\title{UniMiSS: Universal Medical Self-Supervised Learning via Breaking Dimensionality Barrier}


\titlerunning{Universal Medical Self-Supervised Learning}
%
\author{Yutong Xie\inst{1}\orcidID{0000-0002-6644-1250} \and
Jianpeng Zhang\inst{2} \and
Yong Xia\inst{2} \and
Qi Wu\inst{1}\thanks{Corresponding author.}}
\authorrunning{Y. Xie et al.}
%
\institute{The University of Adelaide, Australia \and 
School of Computer Science and Engineering, Northwestern Polytechnical University, China \\
\email{yutong.xie678@gmail.com;qi.wu01@adelaide.edu.au}}
\maketitle

\begin{abstract}
Self-supervised learning (SSL) opens up huge opportunities for medical image analysis that is well known for its lack of annotations. 
However, aggregating massive (unlabeled) 3D medical images like computerized tomography (CT) remains challenging due to its high imaging cost and privacy restrictions. 
In this paper, we advocate bringing a wealth of 2D images like chest X-rays as compensation for the lack of 3D data, aiming to build a universal medical self-supervised representation learning framework, called UniMiSS. 
The following problem is how to break the dimensionality barrier, \ie, making it possible to perform SSL with both 2D and 3D images?
To achieve this, we design a pyramid U-like medical Transformer (MiT).
It is composed of the switchable patch embedding (SPE) module and Transformers. 
The SPE module adaptively switches to either 2D or 3D patch embedding, depending on the input dimension. 
The embedded patches are converted into a sequence regardless of their original dimensions. 
The Transformers model the long-term dependencies in a sequence-to-sequence manner, thus enabling UniMiSS to learn representations from both 2D and 3D images. 
With the MiT as the backbone, we perform the UniMiSS in a self-distillation manner. 
We conduct expensive experiments on six 3D/2D medical image analysis tasks, including segmentation and classification. The results show that the proposed UniMiSS achieves promising performance on various downstream tasks, outperforming the ImageNet pre-training and other advanced SSL counterparts substantially.
Code is available at \def\UrlFont{\rm\small\ttfamily} \url{https://github.com/YtongXie/UniMiSS-code}.

\end{abstract}

\section{Introduction}
\label{sec:intro}

\begin{figure}[!t]
	\begin{center}
		{\includegraphics[width=1.0\linewidth]{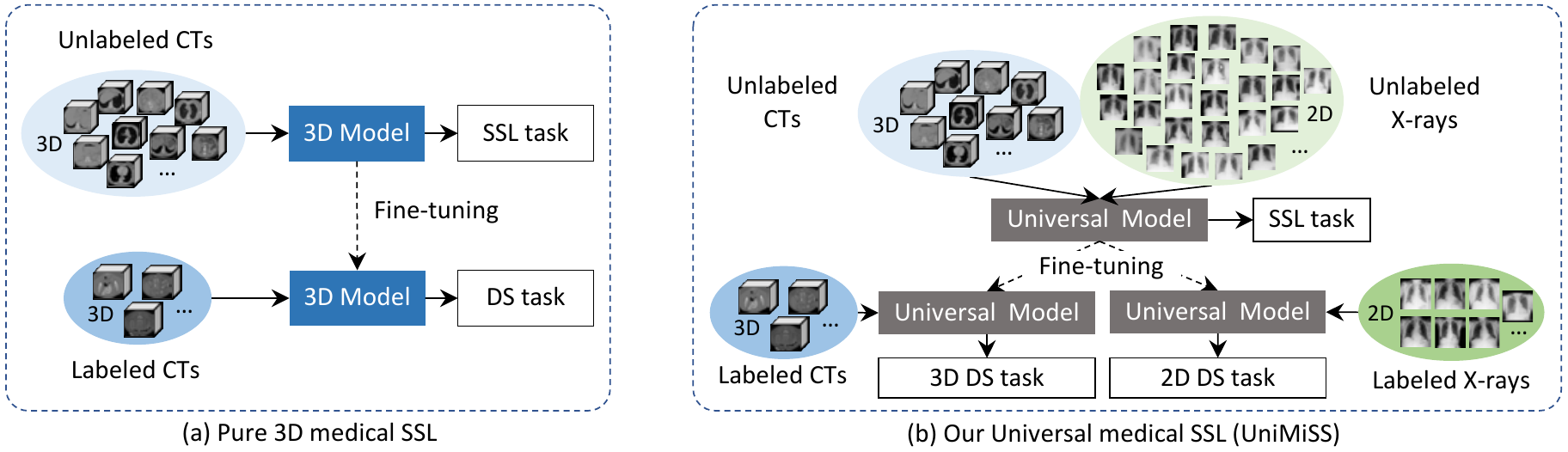}}
	\end{center}
 	\vspace{-0.5cm}
	\caption{
	(a) Pure 3D medical SSL learns representations with only 3D CT scans;
	(b) our proposed UniMiSS brings a wealth of 2D X-rays to offset the lack of 3D data, thus enables the large-scale SSL for better pre-training performance. Besides, the pre-trained model is generic to various downstream (DS) applications, without the restriction on the dimensionality barrier. 
	}	
	\label{fig:fig1}
	\vspace{-0.3cm}
\end{figure}

Medical image analysis, a key process in computer-aided diagnosis, is well known by its lack of labels for training, especially for the 3D task. 
Recent research work suggests that the self-supervised learning (SSL) is promising to ease the annotation cost by making the best of unlabeled data~\cite{GLcontras,SSL_MedIA_resto,MoCoXray,3Dssl_Taleb,PGL,PCRL,3Dssl_zhou,SSL_MedIA_cube}. 
Although setting label free, SSL still heavily relies on the large-scale unlabeled data to explore the feature representations. 
Unfortunately, publicly available 3D medical data is relatively limited due to the high imaging cost and data privacy.
Most of 3D medical datasets just contain a few thousands of cases. 
For example, Zhou~\etal~\cite{PCRL} utilized the LUNA dataset~\cite{LUNA16}, containing about 1000 CT cases, for self-supervised pre-training. 
Such a small data scale may limit the potential of SSL in 3D medical image analysis. 

In comparison to 3D data, it is easy to collect hundreds of thousands of 2D medical images such as X-rays due to its fast imaging speed, low radiation and low cost. 
Accordingly, we advocate to bring a wealth of 2D medical images to the 3D SSL process, aiming at learning strong representations with large-scale images, as shown in Fig.~\ref{fig:fig1}. Comparing to the pure 3D medical SSL, this practice benefits the medical SSL in terms of three significant merits. 
First, 2D data serves as a compensation for the lack of 3D data, enabling the large-scale SSL pre-training. 
Second, there is the anatomy correlation between 2D and 3D images, like chest X-ray and CT. Such an intrinsic relevance may contribute for strong associated representations. 
Third, the pre-trained model is generic enough to be applied to both 3D and 2D downstream tasks. 
To achieve the universal SSL purpose, on the technical side, we need to build a versatile model that is able to process both 2D and 3D images. 
The common practice in medical image analysis is to design 2D convolutional neural networks (CNNs) for 2D images~\cite{Xie_TMI_skin,Zhang_TMI_ano,PCRL} and 3D CNNs for 3D images~\cite{PGL,CoTr,dodnet,PCRL,3Dssl_zhou}, respectively. 
Restricted to the dimensionality barrier, it is almost impossible to design a dimension-free CNN network for this purpose. 

Recent months have witnessed the success of Transformer in computer vision~\cite{ViT}. A vision Transformer usually takes a sequence of image patches, represented by the learned linear embedding, as the input to model the long-term dependencies among the sequence elements.
Owing to the sequence modeling, Transformer can accept the data of any dimensions, including but not limited to 2D images and 3D volumetric data. Therefore, Transformer offers the possibility of breaking the dimensionality barrier and constructing a universal SSL model.

In this paper, we propose a \textbf{Uni}versal \textbf{M}ed\textbf{i}cal \textbf{S}elf-\textbf{S}upervised representation learning framework (UniMiSS) that learns general representations from 2D and 3D unlabeled medical images. 
To achieve this, we design a dimension-free pyramid U-like \textbf{M}ed\textbf{i}cal \textbf{T}ransformer (MiT), which is mainly composed of switchable patch embedding (SPE) module and Transformers. 
The SPE module converts the input images to a sequence by using 2D or 3D patch embedding, depending on the input dimension. 
The Transformer layer processes the embedded tokens in a sequence-to-sequence manner, regardless of their original dimension. 
We perform the self-supervised learning by the self-distillation of student and teacher networks, both of which take the MiT as the backbone. 
The student network learns to predict the output distribution obtained with the momentum teacher network, following the view consistency. 
Moreover, the 3D volumetric image should be identical with their slices due to the same imaging content. 
The volume-slice consistency is adopted as a cross-dimension regularization to boost the representations. 
We conduct the SSL experiments based on 5,022 3D CT volumes, which are augmented by 108,948 2D X-ray images. 
Benefit from the huge augmented 2D data, the proposed UniMiSS achieves the obvious performance improvement on the downstream 3D classification/segmentation tasks. 
Besides, the UniMiSS pre-trained model can be freely applied to 2D downstream tasks, which beats strong competitors like ImageNet pre-training on the downstream 2D medical tasks. 

To summarise, our contributions are three-fold: 
(1) we are the first to augment 3D medical images with the easily accessible unpaired 2D ones for the SSL purpose, aiming at addressing the limitation of 3D data amounts during the SSL process; 
(2) the proposed MiT breaks the dimensionality barrier and enables the joint SSL training with both 2D and 3D images; 
and (3) our UniMiSS pre-training achieves the advanced performance on six downstream tasks, covering the 3D/2D medical image classification/segmentation.

\section{Related Work}
\label{sec:relatedwork}

\subsection{Self-supervised Learning}
SSL has been extensively studied in the literature. According to the pretext tasks, these studies can be broadly categorized into the discriminative methods~\cite{Clustering,SimCLR,BYOL,MoCo,DIM,Transformation,PIPL,Jigsaw,CPC,CMC} and generative methods~\cite{Colorization,Super_resolution,Inpainting,GAN,Decoupling}. The contrastive learning~\cite{SimCLR,BYOL,MoCo,DIM,PIPL,CPC,CMC} has drawn significant research attention and achieved advanced performance on many vision tasks.
Most of the previous work were built on the CNN-based network. 
More recently, Transformer has become an increasingly popular alternative architecture in computer vision. 
There has been a trend towards combining the merits of Transformer and SSL, advancing the self-supervised vision Transformers. 
The seminal work is iGPT~\cite{iGPT}, which follows the masked auto-regressive language modelling to pre-train the self-supervised vision Transformer. 
Besides, some attempts have also been made to pre-train vision Transformers using the contrastive learning~\cite{DINO} or Siamese distillation~\cite{MoCov3}, which outperform the CNN-based SSL approaches, setting a new record on ImageNet. 

The success of SSL in computer vision also benefits to the medical community~\cite{GLcontras,SSL_MedIA_resto,MoCoXray,3Dssl_Taleb,PGL,PCRL,3Dssl_zhou,SSL_MedIA_cube}. Typical attempts include pre-training a CNN by restoring the content of raw images~\cite{SSL_MedIA_resto,3Dssl_Taleb,PCRL,3Dssl_zhou,SSL_MedIA_cube} and tailoring contrastive SSL to medical images~\cite{GLcontras,MoCoXray,3Dssl_Taleb,PGL}.
These efforts constitute an important and timely step forward towards better SSL approaches to medical image analysis. However, they suffer two limitations. 
First, the CNN architecture enables the pre-training on either 2D or 3D medical images, failing to process both of them simultaneously. The resulting representations would be trapped especially for the limited 3D data. 
Consequently, the pre-trained CNN can only be transferred to the dimension-specific downstream task. 
Second, the above SSL approaches capture the spatial context of 3D medical images from either slices~\cite{GLcontras} or volume~\cite{3Dssl_Taleb,3Dssl_zhou,SSL_MedIA_cube}. Few of them consider the inherent consistency relation between volume and its slices.


\subsection{Cross-Domain Training for Medical Imaging}
In the medical context, the cross-domain training usually jointly utilizes two or more datasets acquired at different sites~\cite{CrossD_MICCAI18,CrossD_Liu} or using different imaging modalities~\cite{CrossD_Dou,CrossD_Kli,CrossD_CVPR18} to train a single model that could perform well on diverse datasets.
Karani~\emph{et~al.}~\cite{CrossD_MICCAI18} and Liu~\emph{et~al.}~\cite{CrossD_Liu} trained a single CNN with shared convolutional layers and specific batch normalization layers using the MRI data acquired at each site individually, aiming to tackle the statistical divergence explicitly.
Zhang~\emph{et~al.}~\cite{CrossD_CVPR18} simultaneously learned a volume-to-volume translation using the unpaired CT and MRI data and strong segmentors using synthetic data, which were translated from another modality.
Dou~\emph{et~al.}~\cite{CrossD_Dou} derived a variant of knowledge distillation (KD) to leverage the shared across-modality information between CT and MRI for accurate segmentation of anatomical structures.
Li~\emph{et~al.}~\cite{CrossD_Kli} also introduced KD to the cross-modality analysis of CT and MRI data, but they simultaneously exploited abundant unlabeled data.
These studies are dedicated to analyzing multi-modal/site but fixed dimension (3D) medical images, failing to address the dimensionality barrier in our scenario. 

\section{Methods}
\label{sec:method}

\begin{figure*}[!t]
	\begin{center}
		{\includegraphics[width=1.0\linewidth]{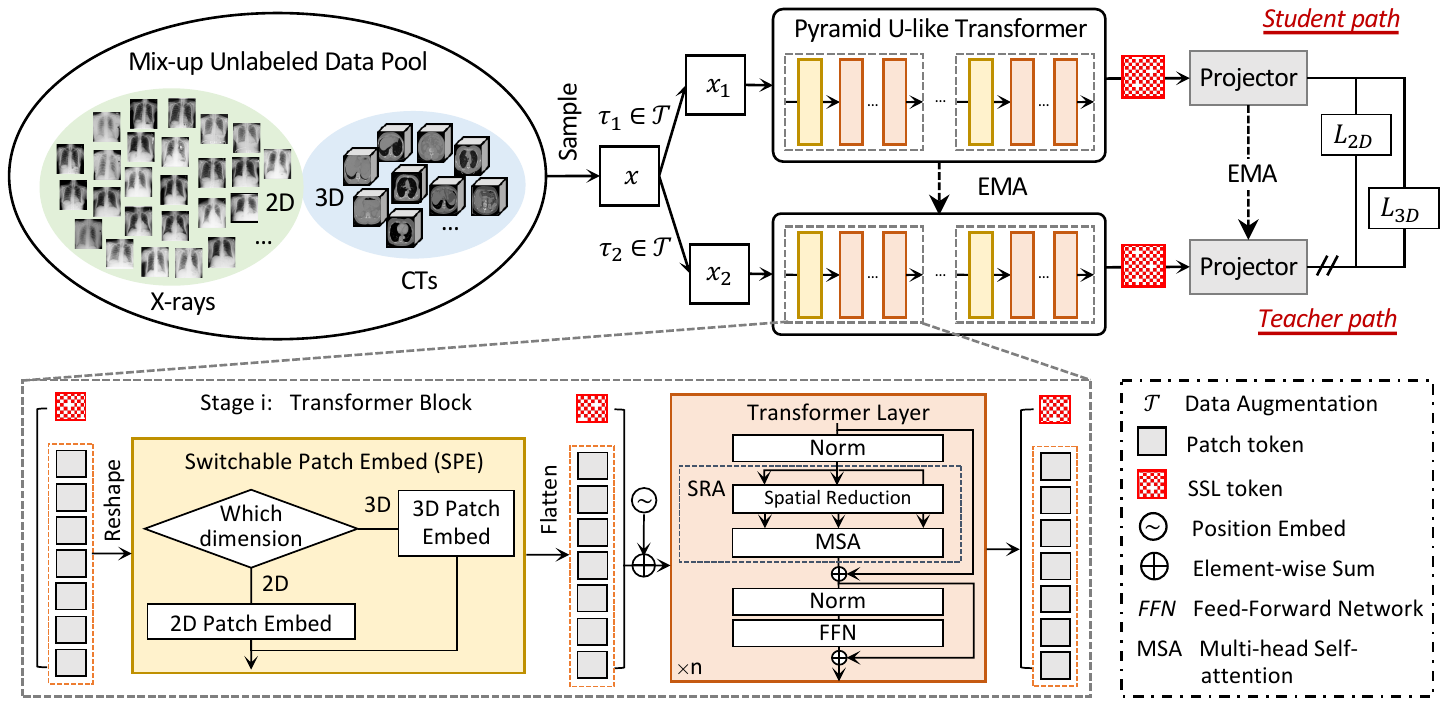}}
	\end{center}
 	\vspace{-0.5cm}
	\caption{Illustration of the proposed UniMiSS framework. It has a dual path architecture, \ie, a student and a teacher. Taking both 2D X-rays and 3D CTs as input, UniMiSS is trained by the self-distillation strategy, \ie, maximizing the agreement of both paths. To break the dimensionality barrier between X-rays and CTs, the MiT network, composed of the switchable patch embedding (SPE) module and Transformers, processes the 3D/2D data in a sequence-to-sequence manner. 
	}	
	\label{fig:fig2}
	\vspace{-0.1cm}
\end{figure*}

\subsection{Overview}

UniMiSS is a universal medical SSL framework that is superior to learn general image representations with large scale mixed 2D and 3D unlabeled medical images. 
Figure~\ref{fig:fig2} illustrates the pipeline of UniMiSS.
Let us denote the mixed 2D and 3D data pool by $\{\mathbb{D}^{2D}, \mathbb{D}^{3D}\}$. 
To enable UniMiSS to process both 2D and 3D medical images, we build the MiT as its backbone, which is mainly constituted by the dimension-adaptive SPE module and Transformer layers. 
We perform the SSL process in the self-distillation manner, and utilize a standard cross-entropy loss to maximize the consistency between the student and teacher outputs. Besides, to get the utmost out of 3D volumetric information, we introduce the volume-slice consistency constraint, which encourages UniMiSS to model the consistency cross dimensions. It is intuitively conducive to learning strong feature representations from the volumetric images. We now delve into the details of this framework.

\subsection{MiT: A Dimension-free Architecture}

Although achieving great success in computer vision, vision Transformer~\cite{ViT} still remains challenging to process high resolution 3D images, due to the high computation cost and memory requirement. 
Inspired by~\cite{PVT}, we design the MiT with a pyramid architecture to process both 2D and 3D images efficiently. 
To break the dimensionality barrier, we propose a simple yet efficient SPE module to adaptively choose the 2D or 3D patch embedding according to the input type. 
MiT has an encoder-decoder architecture that facilitates the various applications, including segmentation and classification. 
We now describe each part of MiT, and more details can be found in Appendix.

\noindent
\textbf{SPE.} 
As shown in Figure~\ref{fig:fig2}, the SPE module plays an important role to obtain the dimension-specific embedding, \ie, using 2D patch embedding operation for 2D inputs and using 3D patch embedding operation for 3D inputs
. Notice that the implementations of SPE in the encoder and decoder are different. The SPE in the encoder refers to a switchable 2D and 3D convolution block with the stride of 2, which reduces the feature resolution. In contrast, the SPE in the decoder is a switchable 2D and 3D transpose convolution block, which increases the feature resolution.

\noindent
\textbf{Encoder-Decoder.} 
The MiT encoder follows a progressive shrinking pyramid Transformer, as done in ~\cite{PVT}. It consists explicitly of four stages, each of which is composed of a SPE module and several stacked Transformers. 
In each stage, the SPE module down-samples the input features and generates the dimension-specific embedded sequence. 
Notably, we append an extra learnable SSL token~\cite{DINO,MoCov3} to the patch embedded sequence. 
The SSL token is similar to the [CLS] token in ViT, which is able to aggregate information from the whole patch embedding tokens via the self-attention.
The resultant sequences, combined with the learnable positional embedding, are inputted into the following Transformers for the long-term dependency modeling. 
Each Transformer layer includes a self-attention module and a feed-forward network (FFN) with two hidden layers.
To enable MiT to process high-resolution images, we follow the spatial-reduction attention (SRA) layer~\cite{PVT}.
Given a query $\bm{q}$, a key $\bm{k}$, and a value $\bm{v}$ as the input, SRA first reduces the spatial resolution of $\bm{k}$ and $\bm{v}$, and then feeds $\bm{q}$, reduced $\bm{k}$, and reduced $\bm{v}$ to a multi-head self-attention (MSA) layer to produce refined features. This process can be formally expressed as follows
\begin{equation}
    SRA(\bm{q},\bm{k},\bm{v}) = MSA(\bm{q}, F(\sigma(R(\bm{k}))), F(\sigma(R(\bm{v})))),
\end{equation}
where $\sigma(\cdot )$ represents a linear projection, \ie, strided 2D or 3D convolution operation, that reduces the feature map resolution, $R(\cdot )$ reshapes the input sequence to a feature map of the original spatial size, and $F(\cdot )$ flattens the input into a 1D sequence.
MiT has a symmetric decoder structure that consists of three stages. In each stage, the input feature map is first up-sampled by the SPE module, and then refined by the stacked Transformer layers.
Besides, we also add skip connections between the encoder and decoder to keep more low-level but high-resolution information.

\subsection{Objective of UniMiSS}

The proposed UniMiSS framework is based on the student-teacher paradigm. 
Each path comprises a MiT network $\mathcal{F}_\theta(\cdot)$ and a projector $\mathcal{P}_\theta(\cdot)$.
$\mathcal{P}_\theta(\cdot)$ is a $n$-layer multi-layer perceptron (MLP) head, $\theta$ represents the parameter set of this path.
The SPE layers switch to perform the 2D patch embedding or 3D patch embedding during the feed-forward computing that is denoted as $\mathcal{F}_\theta(\cdot; 2D)$ and $\mathcal{F}_\theta(\cdot; 3D)$, respectively. 
During the SSL process, we only extract the SSL token from the output of $\mathcal{F}_\theta(\cdot; 2D/3D)$ as the input of the projector. 
Since the Transformer sets the dimension free, our UniMiSS is able to learn image representations from both 2D and 3D unlabeled medical images.

Both of paths share an identical architecture. However, they differ in the following two items. First, the teacher network is formulated as a momentum version of the student network, which updated by an exponential moving average strategy, defined as
\begin{equation}
\mu \leftarrow \lambda\mu+(1-\lambda)\theta, 
\end{equation}
where $\lambda$ increases from 0.996 to 1 using a cosine schedule during training~\cite{DINO}.
Second, a stop-gradient operator is performed to the teacher network to avoid model collapse. 

\noindent
\textbf{Objective for 2D domain data.}
Taking a mini-batch of 2D data $\bm{x}$ for example, we first create two augmented views $\bm{x}_1$ and $\bm{x}_2$ by using the data augmentation module $\mathcal{T}$, and then feed them into the student and teacher networks. The obtained SSL token is inputted into the projector to produce the output vector, denoted as $\bm{f}_1 = \mathcal{P}_\theta(\mathcal{F}_\theta(\bm{x}_1; 2D))$, $\bm{f}_2 = \mathcal{P}_\mu(\mathcal{F}_\mu(\bm{x}_2; 2D))$. 
The objective of UniMiSS is to maximize the consistency between the output vectors obtained with student and teacher networks, formulated by
\begin{equation}
\mathcal{H}(\bm{f}_1, \bm{f}_2)=-\mathrm{softmax}(\frac{\bm{f}_2-\mathcal{C}}{\tau_t}) * \log(\mathrm{softmax}(\frac{\bm{f}_1}{\tau_s})),
\end{equation}
where $\mathcal{C}$ is the centering of teacher outputs, $\tau_t$ and $\tau_s$ are sharpening temperature parameters for student and teacher network. 
The centering operation heartens the model to the uniform distribution while the sharpening has the opposite effect, \ie, encouraging one dimension to dominate. Both of them are jointly used together to avoid model collapse~\cite{DINO}. 
Specifically, the temperature $\tau_t$ is set to a small value in the teacher path for the sharpening purpose.
The center $\mathcal{C}$ is first computed via averaging the teacher's outputs of the min-batch data and then updated with an exponential moving average strategy to aggregate the center across the whole batches, shown as follows
\begin{equation}
\mathcal{C} \leftarrow \omega * \mathcal{C}+(1-\omega )*\widehat{\bm{f}_2}
\end{equation}
where $\omega$ is a rate parameter, and $\widehat{\bm{f}_2}$ refers to the mean of teacher output in a mini-batch. 
We define a symmetrized loss for 2D images as:
\begin{equation}
\mathcal{L}^{\mathrm{2D}}=\mathbb{E}_{\bm{x}\sim \mathbb{D}^{2D}}[\mathcal{H}(\bm{f}_1, \bm{f}_2) + \mathcal{H}(\bm{f}_2, \bm{f}_1)]
\end{equation}

\noindent
\textbf{Objective for 3D domain data.}
In medical domain, 3D volumes can be viewed as the stacking of 2D images along with the inter-slice dimension. The volume data has the inherent consistency to their slices, which inspires us to model the volume-slice consistency for SSL. 
Given a 3D data $\bm{x}$ sampled from the 3D medical dataset, we denote its two augmented views as $\bm{x}_1$ and $\bm{x}_2$, each containing $m$ 2D slices. 
We compute the global volumetric representations by the student and teacher networks in a 3D mode, \ie, $\bm{f}_1 = \mathcal{P}_\theta(\mathcal{F}_\theta(\bm{x}_1; 3D))$, and $\bm{f}_2 = \mathcal{P}_\mu(\mathcal{F}_\mu(\bm{x}_2; 3D))$.
Meanwhile, we stack $m$ slices of each augmented view in a batch, and use them as 2D inputs to calculate the slice-wise representations in a 2D mode, and then treat the average outputs of all slices as the holistic slice representations, \ie, $\bm{f}'_1=\frac{1}{m}\sum_{i=1}^{m}\mathcal{P}_\theta(\mathcal{F}_\theta([\bm{x}_1]^i; 2D))$, and $\bm{f}'_2=\frac{1}{m}\sum_{i=1}^{m}\mathcal{P}_\mu(\mathcal{F}_\mu([\bm{x}_2]^i;2D))$, where $[\bm{x}]^i$ represents the $i$-th slice extracted from the 3D data $\bm{x}$. 
After that, we build the following objective function 
\begin{equation}
\begin{split}
\mathcal{L}^{\mathrm{3D}}=\mathbb{E}_{\bm{x}\sim \mathbb{D}^{3D}}&[\mathcal{H}(\bm{f}_1, \bm{f}_2) + \mathcal{H}(\bm{f}_1, \bm{f}'_2) + \mathcal{H}(\bm{f}'_1, \bm{f}_2) + \mathcal{H}(\bm{f}'_1, \bm{f}'_2) \\
&+\mathcal{H}(\bm{f}_2, \bm{f}_1) + \mathcal{H}(\bm{f}_2, \bm{f}'_1) + \mathcal{H}(\bm{f}'_2, \bm{f}_1) + \mathcal{H}(\bm{f}'_2, \bm{f}'_1)]
\end{split}
\end{equation}
The above objective function encourages to learn the refined consistency with 3D medical data in terms of three aspects, \ie, volume to volume, slice to slice, and volume to slice. 

\begin{figure}[!t]
	\begin{center}
		{\includegraphics[width=0.7\linewidth]{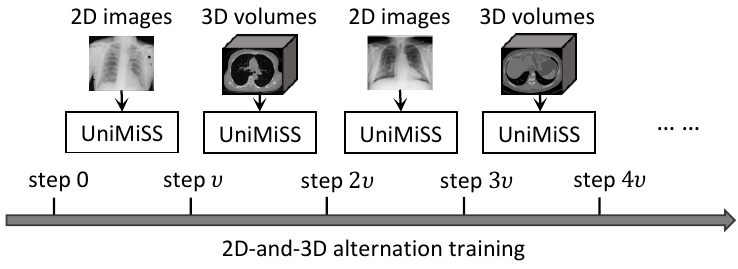}}
	\end{center}
 	\vspace{-0.5cm}
	\caption{Illustration of 2D-and-3D alternation training.}	
	\label{fig:fig3}
	\vspace{-0.1cm}
\end{figure}

We introduce an alternative training scheme to solve this multi-objective optimization problem. 
As shown in Figure~\ref{fig:fig3}, we first sample 2D images to train the UniMiSS from step 0 to step $\upsilon$, and then take turn to sample 3D volumes in the next $\upsilon$ steps. The following training process will continue in a circular manner until the model converges. 
The proposed iterative training scheme has two merits: (1) it bypasses the difficulty of using both 2D and 3D images in the same batch; and (2) it can reduce the instability caused by the distribution discrepancy between 2D and 3D data.

\section{Experiments}
\label{sec:Experiments}

\subsection{Datasets}
\noindent
\textbf{Pre-training datasets.}
We collected 5,022 3D CT scans from five datasets 
(\ie MOTS dataset~\cite{dodnet}, 
LIDC-IDRI dataset~\cite{LIDC}, 
Tianchi dataset~\cite{Tianchi}, 
RibFrac dataset~\cite{RibFrac}, 
TCIACT dataset~\cite{TCIACT}), 
and collected 108,948 2D images from NIH ChestX-ray8 dataset~\cite{Chestxray8} to train UniMiSS in a self-supervised manner. 

\noindent
\textbf{Downstream datasets.}
Table~\ref{tab:tab1} gives the details of six downstream tasks, which can be grouped into (1) 3D downstream: CT-based segmentation (BCV) and classification (RICORD), MRI-based segmentation (CHAOS); (2) 2D downstream: multi-organ segmentation (JSRT) and pneumonia classification (ChestXR), and skin lesion segmentation (ISIC). 
Note that the CHAOS and ISIC datasets are different from the pre-training data in terms of modalities (\ie, 3D CT vs. MRI, 2D X-ray vs. dermoscopy). They are used to evaluate the unseen-modality transferability. 

\begin{table}[t!]
\footnotesize
\caption{Six datasets for the downstream evaluation. Noticed that we used two test sets, \ie offline test set  (off) and online test set (on), for the BCV dataset.}
\label{tab:tab1}
\vspace{-0.5cm}
\begin{center}
\begin{tabular}{c|c|c|c|c}
\hline
\multicolumn{5}{c}{Downstream evaluation datasets}                                 \\ \hline
Name    & Tasks & Modalities & \#Train & \#Test\\ \hline
BCV~\cite{BCV}     & Multi-organ segmentation  & \multirow{2}{*}{3D CT}    & 24 & 6 (off)+20 (on)\\ \cline{1-2} \cline{4-5} 
RICORD~\cite{RICORD}  & COVID-19 screening        &                        & 182 & 45\\ \hline
CHAOS~\cite{CHAOS}   & Abdominal organ segmentation  & 3D MRI                    & 48 & 12\\ \hline
JSRT~\cite{JSRT,SCR}    & Multi-organ segmentation  & \multirow{2}{*}{2D X-ray} & 124 & 123\\ \cline{1-2} \cline{4-5} 
ChestXR~\cite{ChestXR} & Pneumonia  classification &                        & 17,955 & 3,430\\ \hline
ISIC~\cite{ISIC}    & Skin lesion segmentation  & 2D dermoscopy             & 2000 & 600\\ \hline
\end{tabular}
\end{center}
\vspace{-0.4cm}
\end{table}

\subsection{Experimental Details}
\noindent
\textbf{Pre-training setup.}
We set the size of input 2D patches to $224\times224$ and 3D patches to $16\times96\times96$, aiming to weigh the balance between reserving enough information for SSL and reducing computational and spatial complexity to an affordable level. We applied a rich set of data augmentations to create positive views, including colour jittering, Gaussian blur/noise, random crop, zooming, and flip to the inputs for producing two views.
Following~\cite{DINO}, we adopted the AdamW optimizer~\cite{adamw} with a cosine decaying learning rate~\cite{cosine_LR}, a warm-up period of 10 epochs, to train our UniMiSS. We empirically set the initial learning rate to 0.0008, batch size to 192, maximum epochs to 200,
rate parameter $\omega$ to 0.9, and temperature parameter $\tau_t$ and $\tau_s$ to 0.04 and 0.1, respectively. 
It took about 2.5 days to pre-train the UniMiSS using 8 NVIDIA V100 GPUs. We understand this is a big GPU consumption but it saves large amount of time and money to collect 3D medical image data, as we use easily-collected 2D data as the fuel.

\noindent
\textbf{Downstream training setup.}
For the classification, we extracted the pre-trained MiT encoder and appended a FC layer with the output channel as the number of classes for prediction.
For the segmentation, we took the pre-trained MiT encoder and decoder while removing the SSL token, and appended a segmentation head for prediction. 
This head includes a transposed convolutional layer, a Conv-IN-LeakyReLU, and a convolutional layer with the kernel size of 1 and the output channel as the number of classes.
The segmentation performance is measured by the Dice coefficient scores. The classification performance is measured by the area under the receiver operator curve (AUC). 
Note that we randomly split 25\% training samples as a validation set to select the hyper-parameters of UniMiSS in the ablation study. 
The detailed training setups for each downstream task are shown in Appendix.

\begin{table*}[t!]
\caption{Segmentation and classification performance of using different pre-training strategies on the BCV offline test set and RICORD test set.}
\label{tab:tab3}
\vspace{-0.5cm}
\begin{center}
\setlength\tabcolsep{5.5pt}
\begin{tabular}{c|c|l|c|c|c|c|c}
\hline
{\multirow{2}{*}{Methods}}   & {\multirow{2}{*}{Backbone}}  & \multicolumn{3}{c|}{BCV (CT, seg)} & \multicolumn{3}{c}{RICORD (CT, cls)} \\ \cline{3-8} 
&     & 20\%          & 40\%         & 100\%        & 20\%            & 40\%            & 100\%          \\ \hline
Rand. init.     & \multirow{4}{*}{CNN}                                                                & 68.44         & 73.14        & 79.93        & 69.72           & 74.66           & 83.36          \\ \cline{1-1} \cline{3-8}  
MoCo v2~\cite{MoCov2} &                                                                                                     & 71.22& 75.09& 82.05& 73.46& 77.81& 85.46\\ \cline{1-1} \cline{3-8} 
PGL~\cite{PGL}     &                                                                                                     & 72.05& 75.86& 82.57& 73.76& 77.96& 85.61\\ \cline{1-1} \cline{3-8}  
PCRL~\cite{PCRL}    &                                                                                                     & 72.80& 76.05& 82.73& 75.11& 79.01& 86.21\\ \hline
Rand. init.     & \multirow{4}{*}{Transformer}                      & 70.09         & 74.60        & 79.97        & 71.36           & 76.06           & 83.21          \\ \cline{1-1} \cline{3-8} 
MoCo v3~\cite{MoCov3} &                                                                                                              & 74.54& 78.16& 82.02& 75.56& 79.66& 85.16\\ \cline{1-1} \cline{3-8}  
DINO~\cite{DINO}    &                                                                                                             & 75.33& 78.88& 82.61& 76.31& 80.11& 85.91\\ \cline{1-1} \cline{3-8}  
UniMiSS (Ours)    &                                                                                                          & \textbf{77.96}& \textbf{80.97}& \textbf{84.99}& \textbf{78.71}& \textbf{82.96}& \textbf{89.06}\\ \hline
\end{tabular}
\end{center}
\vspace{-0.4cm}
\end{table*}

\subsection{Results on 3D downstream tasks}

\noindent
\textbf{Dimension-specific SSL~\textit{vs.} Cross-dimension SSL.}
In this section, we evaluate the SSL performance on two downstream 3D taks, \ie, multi-organ segmentation (BCV) and COVID-19 screening (RICORD). 
The UniMiSS pre-training is compared with the random initialization (Rand. init.) and five advanced SSL methods, including MoCo v2/v3~\cite{MoCov2,MoCov3}, PGL~\cite{PGL}, PCRL~\cite{PCRL}, and DINO~\cite{DINO}. 
Note that MoCo v2, PGL, and PCRL take the CNN as their encoder backbone, \ie, a 3D ResNet with 50 learnable layers. During the SSL process, MoCo v2 and PGL only pre-train the encoder part, while PCRL additionally pre-trains a decoder by using the reconstruction task. 
Besides, MoCo v3, DINO, and our UniMiSS use the Transformer model as the backbone, which contains both encoder and decoder. 
We employ the U-like PVT as the backbone for MoCo v3 and DINO, which has a similar architecture of MiT but the different patch embedding module. The lack of SPE make them fail to process both 2D and 3D images simultaneously, resulting in the dimension-specific SSL with only 3D data. 
For a fair comparison, all of these SSL methods are pre-trained on the 5,022 unlabeled 3D CT scans. Somewhat differently, the proposed UniMiSS introduces the additional 2D X-rays to the 3D SSL training, benefiting from the universality. 
We make more detailed comparisons between the proposed UniMiSS and other dimension-specific CNN/Transformer SSL methods. 
Table~\ref{tab:tab3} shows the results of three label settings (20\%,40\%, and 100\% label available). 
We summarize this table in the following points: 
(1) The Transformer-based models outperform obviously the CNN-based methods, mainly owing to the SSL pre-training. 
It reflects that the Transformer is a competitive architecture and the SSL pre-training is essential for the Transformer to achieve good performance. 
(2) The proposed UniMiSS is superior to MoCo v3 and DINO. The performance gains over DINO are +2.38\% for segmentation and +3.15\% for classification when 100\% labels are available. 
It proves the effectiveness of using a wealth of 2D medical images to assist the 3D SSL process.
(3) Besides, it is really encouraging to see that the proposed UniMiSS is able to achieve the comparable or even superior performance while less annotations, even a half. Taking BCV for example, UniMiSS with 40\% label achieves 80.97\% segmentation Dice, which is better than the 79.97\% of the random initialized method with 100\% labels.

\noindent
\textbf{Comparisons on the BCV online test set. }
To be more persuasive, we also compared the proposed UniMiSS with other state-of-the-art segmentation methods on the BCV online test set. As listed in Table~\ref{tab:tab_on}, these compared methods include PaNN~\cite{PaNN}, UNETR~\cite{Unetr}, nnUnet~\cite{nnUnet} and DoDnet~\cite{dodnet}. 
Note that the performance records of these competitors come from their original paper. 
It reveals that our UniMiSS, without using any ensemble strategy, still achieves the competitive performance, the best Hausdorff distance (HD) and average mean surface distance (SD), and second highest Dice on the online test set, outperforming the DoDNet with supervised pre-training. When using the coarse-to-fine ensemble strategy like~\cite{nnUnet}, our UniMiSS can obtain the best performance in terms of all metrics.

\begin{table}[t]
\caption{Comparisons on the BCV online test set. }
\label{tab:tab_on}
\vspace{-0.8cm}
\begin{center}
\setlength\tabcolsep{3.5pt}
\begin{tabular}{c|c|c|c|c|c|c}
\hline
Metrics  & PaNN~\cite{PaNN}  & UNETR~\cite{Unetr}   & nnUnet~\cite{nnUnet} & DoDnet~\cite{dodnet} & UniMiSS & UniMiSS \\ \hline
Ensemble  & 5  & 5   & 10 & 5 & 1 & 10 \\ \hline 
Dice & 85.00 & 85.55            & 87.62  & 86.44  & 87.05 & ~\textbf{88.11}        \\ \hline
HD   & 18.47 & \textbackslash{} & \textbackslash{}  & 15.62  & 13.92 & ~\textbf{13.17}        \\ \hline
SD   & 1.45 & \textbackslash{} & \textbackslash{}  & 1.17  & 1.02 & ~\textbf{0.90}        \\ \hline
\end{tabular}
\end{center}
\vspace{-0.3cm}
\end{table}

\noindent
\textbf{Results on 2D downstream tasks.}
Since pre-training on both 2D and 3D medical images, our UniMiSS can be freely applied to 2D downstream tasks.
Table~\ref{tab:tab2} makes the comparisons on the 2D medical image segmentation and classification tasks. 
The compared methods include the Rand. init., ImageNet pre-training (INpre), CNN-based SSL methods (\ie MoCo v2, PGL and PCRL), and Transformer-based SSL methods (\ie MoCo v3 and DINO).
Different from the 3D scenarios, a 2D ResNet-50 is used as the backbone in MoCo v2, PGL, and PCRL. 
MoCo v3, and DINO still take the U-like PVT as the backbone, but modify the patch embedding to adapt for the 2D inputs. 
Here, all compared SSL methods are pre-trained on the same 2D unlabeled medical images.
As for UniMiSS, we directly apply the previous pre-trained model to the 2D tasks, without any modification or further re-training. 
From the results, we can find that
(1) the SSL methods have surpassed INpre in both tasks, revealing that pre-training on a large-scale medical image dataset is more friendly to medical domain downstream tasks than pre-training on natural images; 
(2) although the number of 3D data is much smaller than 2D, \ie, about one in twenty, the UniMiSS pre-training still achieves the performance gain over the pure 2D SSL method, like DINO. 
This may account in part for the inherent correlation between X-rays and CTs. Such a correlation information can be captured by the UniMiSS, thus contributed for the performance gain. 

\begin{table*}[t!]
\caption{Segmentation and classification performance of using different pre-training strategies on two 2D test sets.}
\label{tab:tab2}
\vspace{-0.7cm}
\begin{center}
\setlength\tabcolsep{6pt}
\begin{tabular}{c|c|l|l|l|l|l|l}
\hline
{\multirow{2}{*}{Methods}}   & {\multirow{2}{*}{Backbone}}  & \multicolumn{3}{c|}{JSRT (X-ray, seg)} & \multicolumn{3}{c}{ChestXR (X-ray, cls)} \\ \cline{3-8} 
&    & 20\%        & 40\%        & 100\%      & 20\%         & 40\%         & 100\%       \\ \hline
Rand. init.      & \multirow{5}{*}{CNN}                                                                              & 84.05       & 87.63       & 90.96      & 92.05        & 94.83        & 97.54       \\ \cline{1-1} \cline{3-8}  
INpre~\cite{IN_pretrain} &                                                                                                       & 87.90& 90.01& 91.73& 94.78& 96.26& 98.13\\ \cline{1-1} \cline{3-8} 
MoCo v2~\cite{MoCov2}  &                                                                                                        & 88.65& 91.03& 92.32& 95.22& 96.61& 98.67\\ \cline{1-1} \cline{3-8}  
PGL~\cite{PGL}      &                                                                                                       & 89.01& 91.39& 92.76& 95.56& 96.96& 98.87\\ \cline{1-1} \cline{3-8}  
PCRL~\cite{PCRL}     &                                                                                                     & 89.55& 91.53& 93.07& 95.88& 97.43& 98.99\\ \hline
Rand. init.      & \multirow{4}{*}{Transformer}                            & 85.55       & 88.83       & 91.22      & 92.80        & 95.20        & 97.04       \\ \cline{1-1} \cline{3-8} 
MoCo v3~\cite{MoCov3}  &                                                                                                                & 90.07& 91.75& 92.68& 95.99& 97.33& 98.59\\ \cline{1-1} \cline{3-8} 
DINO~\cite{DINO}     &                                                                                                               & 90.40& 92.16& 93.03& 96.44& 97.69& 98.70\\ \cline{1-1} \cline{3-8} 
UniMiSS    &                                                                                                              & \textbf{91.88}& \textbf{93.15}& \textbf{94.08}& \textbf{97.09}& \textbf{98.14}& \textbf{99.07}\\ \hline
\end{tabular}
\end{center}
\vspace{-0.7cm}
\end{table*}

\begin{figure*}[t!]
	\begin{center}
		{\includegraphics[width=1.0\linewidth]{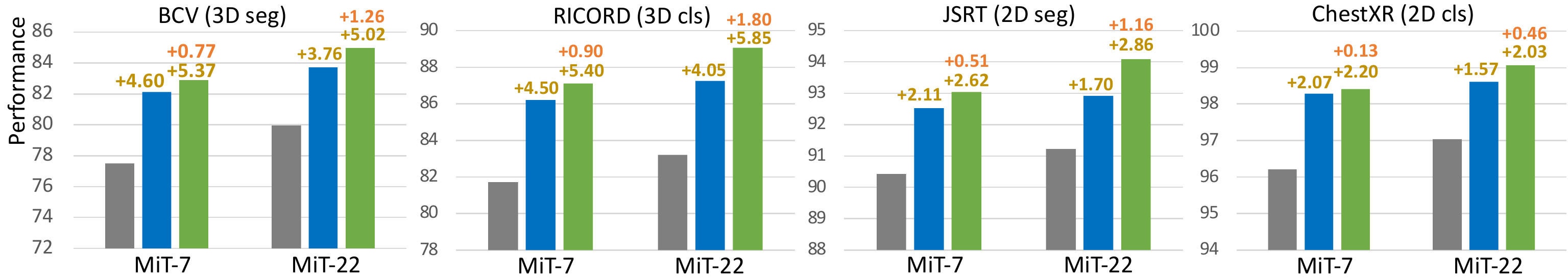}}
	\end{center}
 	\vspace{-0.5cm}
	\caption{Results of MiT with fewer Transformer layers. Here, MiT-7 and MiT-22 denote MiT with 7 and 22 Transformer layers, respectively. 
	{\color{gray}\rule[-0.1em]{0.8em}{0.8em}} Rand. init., {\color{Denim}\rule[-0.1em]{0.8em}{0.8em}} Dimension-specific pre-training, {\color{Green(ryb)}\rule[-0.1em]{0.8em}{0.8em}} UniMiSS. Note that the performance gain with yellow and orange color is computed by comparing to the Rand. init., and dimension-specific pre-training baseline, respectively. }	
	\label{fig:tiny_vs_small}
\end{figure*}

\subsection{Discussions}
\noindent
\textbf{Effectiveness of volume-slice consistency.} 
We design the volume-slice consistency mechanism for learning rich representations with 3D medical images. To evaluate the effectiveness of this mechanism, we pre-trained UniMiSS on 3D medical images with or without using the volume-slice consistency.
Table~\ref{tab:tab5} gives the downstream performance on the validation of two 3D datasets. The proposed volume-slice consistency can substantially improve the 3D segmentation/classification accuracy under different label ratios. The performance gain is at least by 1.42\% on segmentation and by 1.47\% on classification.

\begin{table}[t!]
\caption{Segmentation and classification performance on two 3D validation sets with or without using volume-slice consistency.}
\label{tab:tab5}
\vspace{-0.5cm}
\begin{center}
\setlength\tabcolsep{6pt}
\begin{tabular}{c|c|c|c|c|c|c|c}
\hline
\multicolumn{2}{c|}{Objective for 3D} & \multicolumn{3}{c|}{BCV (seg)} & \multicolumn{3}{c}{RICORD (cls)} \\ \hline
Volume             & Slices            & 20\%          & 40\%         & 100\%        & 20\%            & 40\%            & 100\%          \\ \hline
\checkmark                  &                   & 72.08        & 76.04        & 80.94        & 69.87       & 74.61            & 80.96           \\ \hline
\checkmark                  & \checkmark                 & \textbf{74.56}          & \textbf{77.97}        & \textbf{82.36}        & \textbf{72.46}           & \textbf{76.89}           & \textbf{82.43}          \\ \hline
\end{tabular}
\end{center}
\vspace{-0.6cm}
\end{table}

\noindent
\textbf{Number of iteration interval.} 
The UniMiSS is optimized in a 2D-3D alternation training way, where the iteration interval $\upsilon$ is a critical parameter. 
A smaller $\upsilon$ may lead to insufficient training for each domain. 
A larger $\upsilon$ may make the network forget the information learned from another domain.
To set a suitable $\upsilon$, we pre-trained UniMiSS with various of $\upsilon$, varying from 1 to 3, and fine-tuned them on four downstream tasks. 
Table~\ref{tab:tab6} shows that the pre-trained UniMiSS can achieve the best performance on four downstream tasks when $\upsilon$ equals $2$, and below or above $2$ gives rise to the performance loss. 
Hence, we suggest setting the iteration interval to 2 during the cross-domain pre-training.

\begin{table}[t]
	\caption{Segmentation and classification performance of our UniMiSS with different iteration intervals on the validation sets.}
	\label{tab:tab6}
    \vspace{-0.8cm}
	\begin{center}
	\begin{tabular}{c|c|c|c|c}
	\hline
	Iteration interval	   & BCV (3D seg)          & RICORD (3D cls)         & JSRT (2D seg)           & ChestXR (2D cls)           \\ \hline
	1                          & 82.70        & 82.95       & 92.33         & 96.65         \\ \hline
	2                          & \textbf{83.56}        & \textbf{84.26}       & \textbf{93.48}         & \textbf{97.57}         \\ \hline
	3                          & 83.28        & 83.65       & 93.12         & 97.16          \\ \hline
	\end{tabular}
	\end{center}
\vspace{-0.3cm}
\end{table}

\noindent
\textbf{MiT with different Transformer scales.} 
Transformer is the dominant component in the MiT backbone. 
We investigate the effect of Transformer scales in MiT. Specifically, we compare a MiT with 22 Transformer layers (MiT-22) and another with seven layers (MiT-7).
The segmentation and classification performance is given in Figure~\ref{fig:tiny_vs_small}, from which three conclusions can be drawn:
(1) increasing the Transformer layers boosts the performance of MiT in all downstream tasks;
(2) as MiT goes deeper, the performance gain of the dimension-specific pre-training over the random initialization becomes smaller, while the performance gain of our UniMiSS with cross-dimension pre-training is basically impregnable; and
(3) the superiority of our UniMiSS pre-training over the dimension-specific pre-training is more evident with the increase of Transformer layers.

\noindent
\textbf{Transferability on unseen modality data.} 
In the above experiments, the pre-training and downstream tasks are all based on CT and X-ray images. To evaluate the transferability of UniMiSS on unseen modalities, we further tested the MoCo v3, DINO and our UniMiSS on the CHAOS dataset (MRI scans) and ISIC dataset (dermoscopic images). 
The results in Table~\ref{tab:tab7} show that UniMiSS can consistently 
improve at least 2.98\% on the CHAOS dataset, and 1.60\% on the ISIC dataset, compared to the random initialization. 
It demonstrates that UniMiSS has a great potential in transferring learned knowledge to the unseen modality. Besides, our UniMiSS also outperforms two popular Transformer-based SSL methods on both CHAOS and ISIC datasets.

\begin{table}[t!]
\caption{Segmentation performance of using the random initialization and three pre-training strategies on CHAOS dataset (unseen MRI scans) and ISIC dataset (unseen dermoscopic images).}
\label{tab:tab7}
\vspace{-0.5cm}
\begin{center}
\setlength\tabcolsep{6pt}
\begin{tabular}{c|c|c|c|c|c|c}
\hline
\multirow{3}{*}{Methods} & \multicolumn{6}{c}{Downstream data}                              \\ \cline{2-7} 
                         & \multicolumn{3}{c|}{2D dermoscopic} & \multicolumn{3}{c}{3D MRI} \\ \cline{2-7} 
                         & 20\%       & 40\%       & 100\%     & 20\%    & 40\%    & 100\%   \\ \hline
Rand. init.              & 76.31      & 79.92      & 85.07     & 73.28   & 83.64   & 88.38   \\ \hline
MoCo v3                  & 78.66      & 81.46      & 86.04     & 78.42   & 87.22   & 89.83   \\ \hline
DINO                     & 79.11      & 81.89      & 86.21     & 79.16   & 87.79   & 90.52   \\ \hline
UniMiSS                     & \textbf{79.78}       & \textbf{82.33}       & \textbf{86.67}     & \textbf{80.50}   & \textbf{88.58}    & \textbf{91.36}   \\ \hline
\end{tabular}
\end{center}
\vspace{-0.4cm}
\end{table}

\noindent
\textbf{Necessity of SPE.} 
Without the SPE module, a straightforward solution is to flatten the pixels or patches and then use a linear layer for the embedding. 
Such a crude flattening operation suffers the high computation complexity and memory requirements, especially for 3D images. 
Accordingly, SPE is an indispensable part of UniMiSS, which enables to
(1) adaptively choose the patch embedding according to the input type; and
(2) lessen the length of the sequence to reduce computation cost when the network goes deep.

\begin{figure*}[t!]
	\begin{center}
		{\includegraphics[width=0.98\linewidth]{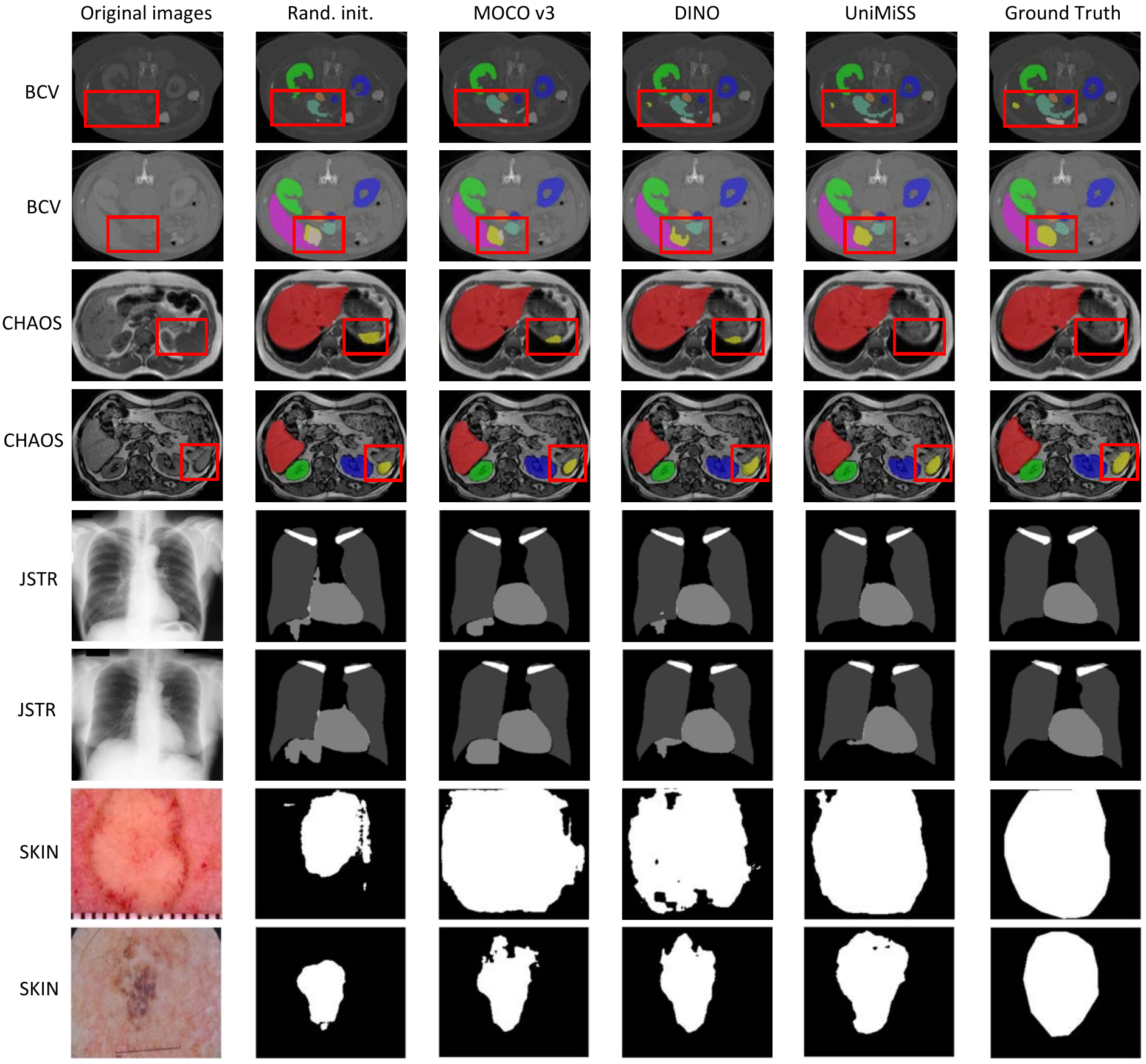}}
	\end{center}
 	\vspace{-0.6cm}
	\caption{Visualization of segmentation results of 8 cases selected from four datasets. The regions in red rectangles indicate our superiority. Our UniMiSS pre-training results in more accurate results than random initialization and other two pre-training strategies. Each type of organs and tumors in single dataset is denoted by a unique color.}	
	\label{fig:fig4}
	\vspace{-0.3cm}
\end{figure*}

\noindent
\textbf{Visualization of Segmentation Results.} 
In Figure~\ref{fig:fig4}, we visualize the segmentation results obtained by the segmentation network, which is initialized (1) randomly, (2) by using the pre-trained MoCo v3~\cite{MoCov3}, (3) by using the pre-trained DINO~\cite{DINO}, or (4) by using our pre-trained UniMiSS.
It shows that our UniMiSS pre-training produces the higher-quality segmentation results, which are more similar to the ground truth, than MoCo v3 and DINO pre-training. Compared to other competitors, UniMiSS pre-training is superior to process challenging cases, like small objects or blurry boundaries. 

\section{Conclusion}
We propose a simple yet effective UniMiSS framework, which introduces a wealth of 2D medical images (\ie X-rays) to the 3D SSL, aiming at making up for the lack of 3D data (\ie CT scans). 
To break the difficulty of dimensionality barrier, we design the MiT as a bridge to connect different dimensions.
In the future, we will extend our UniMiSS to deal with more dimensions (\eg clinic text or genetic data).


\noindent
\textbf{Acknowledgement}
Jianpeng Zhang and Yong Xia were supported by National Natural Science Foundation of China under Grants 62171377. 
Qi Wu was funded by ARC DE190100539.

\bibliographystyle{splncs04}\appendix
\bibliography{egbib}

\begin{thebibliography}{10}
\providecommand{\url}[1]{\texttt{#1}}
\providecommand{\urlprefix}{URL }
\providecommand{\doi}[1]{https://doi.org/#1}

\bibitem{BCV}
Multi-atlas labeling beyond the cranial vault - workshop and
  challenge.~\url{https://www.synapse.org/\#!Synapse:syn3193805/wiki/217789}

\bibitem{Tianchi}
Tianchi
  dataset.~\url{https://tianchi.aliyun.com/competition/entrance/231601/information?from=oldUrl}

\bibitem{ChestXR}
Akhloufi, M.A., Chetoui, M.: {Chest XR COVID-19 detection}.
  \url{https://cxr-covid19.grand-challenge.org/} (August 2021), online;
  accessed September 2021

\bibitem{TCIACT}
An, P., Xu, S., Harmon, S., Turkbey, E., Sanford, T., Amalou, A., Kassin, M.,
  Varble, N., Blain, M., Anderson, V., et~al.: Ct images in covid-19 [data
  set]~\url{https://doi.org/10.7937/TCIA.2020.GQRY-NC81}. The Cancer Imaging
  Archive  (2020)

\bibitem{LIDC}
Armato~III, S.G., McLennan, G., Bidaut, L., McNitt-Gray, M.F., Meyer, C.R.,
  Reeves, A.P., Zhao, B., Aberle, D.R., Henschke, C.I., Hoffman, E.A., et~al.:
  The lung image database consortium (lidc) and image database resource
  initiative (idri): a completed reference database of lung nodules on ct
  scans. Medical physics  \textbf{38}(2),  915--931 (2011)

\bibitem{Clustering}
Caron, M., Bojanowski, P., Joulin, A., Douze, M.: Deep clustering for
  unsupervised learning of visual features. In: ECCV. pp. 132--149 (2018)

\bibitem{DINO}
Caron, M., Touvron, H., Misra, I., J\'egou, H., Mairal, J., Bojanowski, P.,
  Joulin, A.: Emerging properties in self-supervised vision transformers. In:
  ICCV (2021)

\bibitem{GLcontras}
Chaitanya, K., Erdil, E., Karani, N., Konukoglu, E.: Contrastive learning of
  global and local features for medical image segmentation with limited
  annotations. In: NeurIPS. vol.~33 (2020)

\bibitem{SSL_MedIA_resto}
Chen, L., Bentley, P., Mori, K., Misawa, K., Fujiwara, M., Rueckert, D.:
  Self-supervised learning for medical image analysis using image context
  restoration. Medical Image Analysis  \textbf{58},  101539 (2019)

\bibitem{iGPT}
Chen, M., Radford, A., Child, R., Wu, J., Jun, H., Luan, D., Sutskever, I.:
  Generative pretraining from pixels. In: ICML. pp. 1691--1703 (2020)

\bibitem{SimCLR}
Chen, T., Kornblith, S., Norouzi, M., Hinton, G.: A simple framework for
  contrastive learning of visual representations. In: ICML (2020)

\bibitem{MoCov2}
Chen, X., Fan, H., Girshick, R., He, K.: Improved baselines with momentum
  contrastive learning. arXiv preprint arXiv:2003.04297  (2020)

\bibitem{MoCov3}
Chen*, X., Xie*, S., He, K.: An empirical study of training self-supervised
  vision transformers. In: ICCV (2021)

\bibitem{ISIC}
Codella, N.C., Gutman, D., Celebi, M.E., Helba, B., Marchetti, M.A., Dusza,
  S.W., Kalloo, A., Liopyris, K., Mishra, N., Kittler, H., et~al.: Skin lesion
  analysis toward melanoma detection: A challenge at the 2017 international
  symposium on biomedical imaging (isbi), hosted by the international skin
  imaging collaboration (isic). In: ISBI). pp. 168--172. IEEE (2018)

\bibitem{ViT}
Dosovitskiy, A., Beyer, L., Kolesnikov, A., Weissenborn, D., Zhai, X.,
  Unterthiner, T., Dehghani, M., Minderer, M., Heigold, G., Gelly, S., et~al.:
  An image is worth 16x16 words: Transformers for image recognition at scale.
  In: ICLR (2021)

\bibitem{CrossD_Dou}
Dou, Q., Liu, Q., Heng, P.A., Glocker, B.: Unpaired multi-modal segmentation
  via knowledge distillation. IEEE Transactions on Medical Imaging
  \textbf{39}(7),  2415--2425 (2020)

\bibitem{BYOL}
Grill, J.B., Strub, F., Altch{\'e}, F., Tallec, C., Richemond, P., Buchatskaya,
  E., Doersch, C., Avila~Pires, B., Guo, Z., Gheshlaghi~Azar, M., et~al.:
  Bootstrap your own latent-a new approach to self-supervised learning. In:
  NeurIPS (2020)

\bibitem{Unetr}
Hatamizadeh, A., Tang, Y., Nath, V., Yang, D., Myronenko, A., Landman, B.,
  Roth, H.R., Xu, D.: Unetr: Transformers for 3d medical image segmentation.
  In: Proceedings of the IEEE/CVF Winter Conference on Applications of Computer
  Vision. pp. 574--584 (2022)

\bibitem{MoCo}
He, K., Fan, H., Wu, Y., Xie, S., Girshick, R.: Momentum contrast for
  unsupervised visual representation learning. In: CVPR. pp. 9729--9738 (2020)

\bibitem{IN_pretrain}
He, K., Zhang, X., Ren, S., Sun, J.: Deep residual learning for image
  recognition. In: Proceedings of the IEEE conference on computer vision and
  pattern recognition. pp. 770--778 (2016)

\bibitem{DIM}
Hjelm, R.D., Fedorov, A., Lavoie-Marchildon, S., Grewal, K., Bachman, P.,
  Trischler, A., Bengio, Y.: Learning deep representations by mutual
  information estimation and maximization. In: ICLR (2019)

\bibitem{nnUnet}
Isensee, F., Jaeger, P.F., Kohl, S.A., Petersen, J., Maier-Hein, K.H.: nnu-net:
  a self-configuring method for deep learning-based biomedical image
  segmentation. Nature methods  \textbf{18}(2),  203--211 (2021)

\bibitem{RibFrac}
Jin, L., Yang, J., Kuang, K., Ni, B., Gao, Y., Sun, Y., Gao, P., Ma, W., Tan,
  M., Kang, H., Chen, J., Li, M.: Deep-learning-assisted detection and
  segmentation of rib fractures from ct scans: Development and validation of
  fracnet. EBioMedicine  (2020)

\bibitem{CrossD_MICCAI18}
Karani, N., Chaitanya, K., Baumgartner, C., Konukoglu, E.: A lifelong learning
  approach to brain mr segmentation across scanners and protocols. In: MICCAI.
  pp. 476--484. Springer (2018)

\bibitem{CHAOS}
Kavur, A.E., Selver, M.A., Dicle, O., Barış, M., Gezer, N.S.: {CHAOS -
  Combined (CT-MR) Healthy Abdominal Organ Segmentation Challenge Data}  (Apr
  2019). \doi{10.5281/zenodo.3362844},
  \url{https://doi.org/10.5281/zenodo.3362844}

\bibitem{Colorization}
Larsson, G., Maire, M., Shakhnarovich, G.: Colorization as a proxy task for
  visual understanding. In: CVPR. pp. 6874--6883 (2017)

\bibitem{Super_resolution}
Ledig, C., Theis, L., Husz{\'a}r, F., Caballero, J., Cunningham, A., Acosta,
  A., Aitken, A., Tejani, A., Totz, J., Wang, Z., et~al.: Photo-realistic
  single image super-resolution using a generative adversarial network. In:
  CVPR. pp. 4681--4690 (2017)

\bibitem{Transformation}
Lee, H., Hwang, S.J., Shin, J.: Self-supervised label augmentation via input
  transformations. In: ICML (2020)

\bibitem{CrossD_Kli}
Li, K., Wang, S., Yu, L., Heng, P.A.: Dual-teacher++: Exploiting intra-domain
  and inter-domain knowledge with reliable transfer for cardiac segmentation.
  IEEE Transactions on Medical Imaging  (2020)

\bibitem{CrossD_Liu}
Liu, Q., Dou, Q., Yu, L., Heng, P.A.: Ms-net: multi-site network for improving
  prostate segmentation with heterogeneous mri data. IEEE Transactions on
  Medical Imaging  \textbf{39}(9),  2713--2724 (2020)

\bibitem{cosine_LR}
Loshchilov, I., Hutter, F.: Sgdr: Stochastic gradient descent with warm
  restarts. In: ICLR (2017)

\bibitem{adamw}
Loshchilov, I., Hutter, F.: Fixing weight decay regularization in adam  (2018)

\bibitem{PIPL}
Misra, I., Maaten, L.v.d.: Self-supervised learning of pretext-invariant
  representations. In: CVPR. pp. 6707--6717 (2020)

\bibitem{Jigsaw}
Noroozi, M., Favaro, P.: Unsupervised learning of visual representations by
  solving jigsaw puzzles. In: ECCV. pp. 69--84. Springer (2016)

\bibitem{CPC}
Oord, A.v.d., Li, Y., Vinyals, O.: Representation learning with contrastive
  predictive coding. arXiv preprint arXiv:1807.03748  (2018)

\bibitem{Inpainting}
Pathak, D., Krahenbuhl, P., Donahue, J., Darrell, T., Efros, A.A.: Context
  encoders: Feature learning by inpainting. In: CVPR. pp. 2536--2544 (2016)

\bibitem{GAN}
Radford, A., Metz, L., Chintala, S.: Unsupervised representation learning with
  deep convolutional generative adversarial networks. arXiv preprint
  arXiv:1511.06434  (2015)

\bibitem{LUNA16}
Setio, A.A.A., Traverso, A., De~Bel, T., Berens, M.S., Van Den~Bogaard, C.,
  Cerello, P., Chen, H., Dou, Q., Fantacci, M.E., Geurts, B., et~al.:
  Validation, comparison, and combination of algorithms for automatic detection
  of pulmonary nodules in computed tomography images: the luna16 challenge.
  Medical Image Analysis  \textbf{42},  1--13 (2017)

\bibitem{JSRT}
Shiraishi, J., Katsuragawa, S., Ikezoe, J., Matsumoto, T., Kobayashi, T.,
  Komatsu, K.i., Matsui, M., Fujita, H., Kodera, Y., Doi, K.: Development of a
  digital image database for chest radiographs with and without a lung nodule:
  receiver operating characteristic analysis of radiologists' detection of
  pulmonary nodules~\url{http://db.jsrt.or.jp/eng.php}. American Journal of
  Roentgenology  \textbf{174}(1),  71--74 (2000)

\bibitem{MoCoXray}
Sowrirajan, H., Yang, J., Ng, A.Y., Rajpurkar, P.: Moco pretraining improves
  representation and transferability of chest x-ray models. In: MIDL. pp.
  728--744. PMLR (2021)

\bibitem{3Dssl_Taleb}
Taleb, A., Loetzsch, W., Danz, N., Severin, J., Gaertner, T., Bergner, B.,
  Lippert, C.: 3d self-supervised methods for medical imaging. In: NeurIPS.
  vol.~33, pp. 18158--18172 (2020)

\bibitem{CMC}
Tian, Y., Krishnan, D., Isola, P.: Contrastive multiview coding  (2020)

\bibitem{RICORD}
Tsai, E.B., Simpson, S., Lungren, M.P., Hershman, M., Roshkovan, L., Colak, E.,
  Erickson, B.J., Shih, G., Stein, A., Kalpathy-Cramer, J., et~al.: The rsna
  international covid-19 open radiology database (ricord). Radiology
  \textbf{299}(1),  E204--E213 (2021)

\bibitem{SCR}
Van~Ginneken, B., Stegmann, M.B., Loog, M.: Segmentation of anatomical
  structures in chest radiographs using supervised methods: a comparative study
  on a public
  database~\url{https://www.isi.uu.nl/Research/Databases/SCR/index.php}.
  Medical Image Analysis  \textbf{10}(1),  19--40 (2006)

\bibitem{PVT}
Wang, W., Xie, E., Li, X., Fan, D.P., Song, K., Liang, D., Lu, T., Luo, P.,
  Shao, L.: Pyramid vision transformer: A versatile backbone for dense
  prediction without convolutions. In: ICCV (2021)

\bibitem{Chestxray8}
Wang, X., Peng, Y., Lu, L., Lu, Z., Bagheri, M., Summers, R.M.: Chestx-ray8:
  Hospital-scale chest x-ray database and benchmarks on weakly-supervised
  classification and localization of common thorax diseases. In: CVPR. pp.
  2097--2106 (2017)

\bibitem{PGL}
Xie, Y., Zhang, J., Liao, Z., Xia, Y., Shen, C.: Pgl: Prior-guided local
  self-supervised learning for 3d medical image segmentation. arXiv preprint
  arXiv:2011.12640  (2020)

\bibitem{CoTr}
Xie, Y., Zhang, J., Shen, C., Xia, Y.: Cotr: Efficiently bridging cnn and
  transformer for 3d medical image segmentation. In: MICCAI. pp. 171--180.
  Springer (2021)

\bibitem{Xie_TMI_skin}
Xie, Y., Zhang, J., Xia, Y., Shen, C.: A mutual bootstrapping model for
  automated skin lesion segmentation and classification. IEEE Transactions on
  Medical Imaging  \textbf{39}(7),  2482--2493 (2020)

\bibitem{MBDCNN}
Xie, Y., Zhang, J., Xia, Y., Shen, C.: A mutual bootstrapping model for
  automated skin lesion segmentation and classification. IEEE Transactions on
  Medical Imaging  \textbf{39}(7),  2482--2493 (2020)

\bibitem{Zhang_TMI_ano}
Zhang, J., Xie, Y., Pang, G., Liao, Z., Verjans, J., Li, W., Sun, Z., He, J.,
  Li, Y., Shen, C., et~al.: Viral pneumonia screening on chest x-rays using
  confidence-aware anomaly detection. IEEE Transactions on Medical Imaging
  \textbf{40}(3),  879--890 (2020)

\bibitem{dodnet}
Zhang, J., Xie, Y., Xia, Y., Shen, C.: Dodnet: Learning to segment multi-organ
  and tumors from multiple partially labeled datasets. In: CVPR. pp. 1195--1204
  (2021)

\bibitem{Decoupling}
Zhang, R., Isola, P., Efros, A.A.: Split-brain autoencoders: Unsupervised
  learning by cross-channel prediction. In: CVPR. pp. 1058--1067 (2017)

\bibitem{CrossD_CVPR18}
Zhang, Z., Yang, L., Zheng, Y.: Translating and segmenting multimodal medical
  volumes with cycle-and shape-consistency generative adversarial network. In:
  CVPR. pp. 9242--9251 (2018)

\bibitem{PCRL}
Zhou, H.Y., Lu, C., Yang, S., Han, X., Yu, Y.: Preservational learning improves
  self-supervised medical image models by reconstructing diverse contexts. In:
  ICCV. pp. 3499--3509 (2021)

\bibitem{PaNN}
Zhou, Y., Li, Z., Bai, S., Wang, C., Chen, X., Han, M., Fishman, E., Yuille,
  A.L.: Prior-aware neural network for partially-supervised multi-organ
  segmentation. In: ICCV. pp. 10672--10681 (2019)

\bibitem{3Dssl_zhou}
Zhou, Z., Sodha, V., Pang, J., Gotway, M.B., Liang, J.: Models genesis. Medical
  Image Analysis  \textbf{67},  101840 (2021)

\bibitem{SSL_MedIA_cube}
Zhu, J., Li, Y., Hu, Y., Ma, K., Zhou, S.K., Zheng, Y.: Rubik’s cube+: A
  self-supervised feature learning framework for 3d medical image analysis.
  Medical Image Analysis  \textbf{64},  101746 (2020)

\end{thebibliography}

\newpage

\section*{\centering{Appendix}}
\section{Overview}
In this document, we provide more discussions and experimental details to supplement the main submission. 
We first continue to discuss the necessity of switchable patch embedding (SPE) module (Section~\ref{Sec.Necessity_SPE}).
We then give more details for the downstream tasks, including the implementation details and architectures (Section~\ref{Sec.Details_Downstream}). 
Finally, we provide an intuitive explanation of the proposed volume-slice consistency mechanism (Section~\ref{Sec.Details_consist}).

\section{Necessity of SPE (Cont.)}
\label{Sec.Necessity_SPE}
To further explain the necessity of the SPE module, we compared the pyramid U-like medical Transformer (MiT) to two variants without a SPE module.
For the variant 1, we directly flatten the 2D/3D images to a sequence based on the pixels/voxels level and then use a linear layer for the embedding. 
Such a crude flattening operation suffers the very high computation complexity and memory requirements, especially for 3D images. 
Thus, it is hard to perform the variant 1 for quantitative comparisons.
For the variant 2, we perform a naive embedding strategy to reduce the complexity. We first down-sample (for encoder)/up-sample (for decoder) the 2D/3D images by using a parameter free interpolation, then flatten them into a sequence based on the pixels/voxels level, and finally use a linear layer for both 2D and 3D embedding. 
The results in Table~\ref{tab:tab_apex1} show that MiT with the SPE module is significantly superior to the naive embedding strategy (\ie variant 2) whenever with or without using the pre-training. 
It suggests that our SPE is better than the parameter-free interpolation and linear layer. 
The reason may be that the strided convolution with a large kernel is able to model the local continuity of 2D/3D images, which cannot be implemented by the linear layer. 

\begin{table}[h!]
	\scriptsize
	\caption{Segmentation performance of MiT and its two variants without a SPE module on BCV offline test set (3D CT).}
	\label{tab:tab_apex1}
	\vspace{-0.5cm}
	\begin{center}
		\renewcommand\arraystretch{1.1}
		\setlength\tabcolsep{7pt}
		\begin{tabular}{cc|c|c}
			\hline
			\multicolumn{2}{c|}{Methods}                                  & SPE & Dice         \\ \hline
			\multicolumn{1}{c|}{\multirow{3}{*}{Random initialization}} & Variant 1 & No  & unaffordable \\ \cline{2-4} 
			\multicolumn{1}{c|}{}                             & Variant 2 & No  & 73.31        \\ \cline{2-4} 
			\multicolumn{1}{c|}{}                             & Ours       & Yes & 79.93        \\ \hline
			\multicolumn{1}{c|}{\multirow{3}{*}{UniMiSS pre-training}}     & Variant 1 & No  & unaffordable \\ \cline{2-4} 
			\multicolumn{1}{c|}{}                             & Variant 2 & No  & 76.65        \\ \cline{2-4} 
			\multicolumn{1}{c|}{}                             & Ours       & Yes & 84.99        \\ \hline
		\end{tabular}
	\end{center}
	\vspace{-0.2cm}
\end{table}

\section{Downstream Tasks}
\label{Sec.Details_Downstream}

\subsection{Implementation Details}
In Table~\ref{tab:tab_apex2}, we provide the implementation details of six downstream datasets, including the task type, modality, number of training and test cases, loss function, patch size, batch size, optimizer, learning rate, and maximum iterations. 
Note that we randomly split 25\% training scans as a validation set to select the hyper-parameters of UniMiSS in the ablation study. 
We use the online data augmentation to alleviate the over-fitting of UniMiSS on training data. We augment 2D images via random cropping and zooming, random rotation, shear, shift, and horizontal/vertical flip. As for 3D images, we perform random rotation, scaling, flipping, adding white Gaussian noise, Gaussian blurring, adjusting rightness and contrast, simulation of low resolution, and Gamma transformation~\cite{nnUnet}. 
All the downstream experiments were performed on a NVIDIA GTX 2080Ti GPU.

\begin{table*}[t!]
	\scriptsize
	\caption{Implementation details of downstream tasks. Seg: Segmentation; Cls: Classification; CE: Cross-entropy loss; off: offline test set; on: online test set.}
	\label{tab:tab_apex2}
	\vspace{-0.6cm}
	\begin{center}
		\renewcommand\arraystretch{1.25}
		\setlength\tabcolsep{2.5pt}
		\begin{tabular}{l|c|c|c|c|c|c}
			\hline
			Dataset & BCV          & RICORD         & JSRT         & ChestXR        & CHAOS        & ISIC           \\ \hline
			Task              & Seg & Cls & Seg & Cls & Seg & Seg   \\ \hline
			Modality          & 3D CT        & 3D CT          & 2D X-ray     & 2D X-ray       & 3D MRI       & 2D Dermoscopic \\ \hline
			Training data     & 24           & 182            & 124          & 17,955         & 16           & 2000           \\ \hline
			Test data         & 6 (off)+20 (on)            & 45             & 123          & 3,430          & 4            & 600            \\ \hline
			Loss              & Dice+CE~\cite{nnUnet}       & CE             & Dice+CE      & CE             & Dice+CE~\cite{nnUnet}      & Hybrid loss~\cite{MBDCNN}    \\ \hline
			Patch size        & $48\times192^2$   & $64\times128^2$     & $224^2$      & $224^2$        & $48\times192\times256$   & $224^2$        \\ \hline
			Augmentation & \checkmark & \checkmark & \checkmark & \checkmark & \checkmark & \checkmark \\ \hline
			Optimizer         & AdamW        & AdamW          & AdamW        & AdamW          & AdamW        & AdamW          \\ \hline
			Learning rate     & 0.0001       & 0.00001        & 0.0001       & 0.0001         & 0.0001       & 0.0001         \\ \hline
			Batch size        & 2            & 8              & 32           & 32             & 2            & 16             \\ \hline
			Iterations   & 25,000       & 14,000         & 10,000       & 17,000         & 50,000       & 37,500         \\ \hline
		\end{tabular}
	\end{center}
	\vspace{-0.4cm}
\end{table*}

\subsection{Architectures of MiT and ResUnet}
\begin{figure}[!h]
	\begin{center}
		{\includegraphics[width=1.0\linewidth]{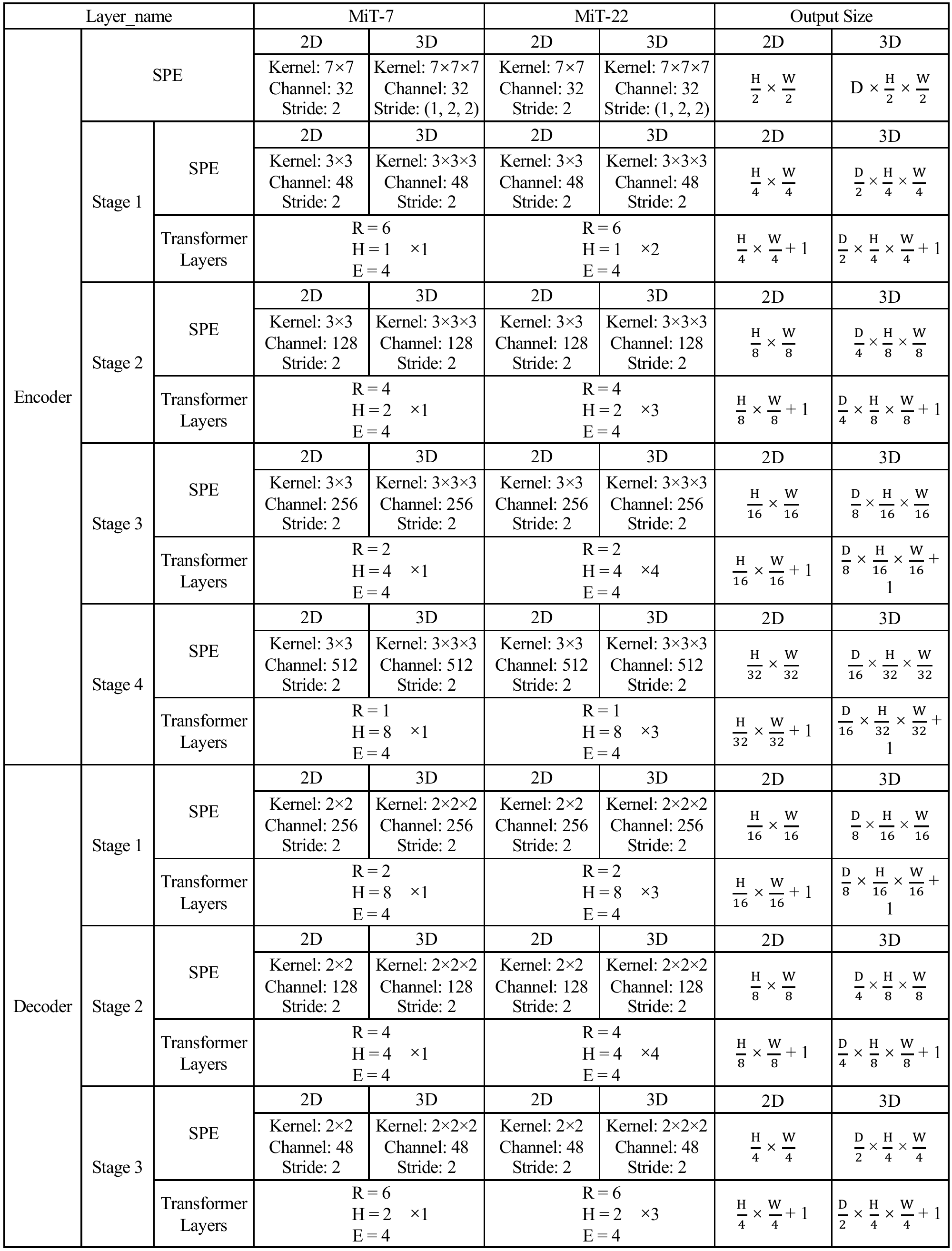}}
	\end{center}
	\vspace{-0.5cm}
	\caption{Detailed settings of MiT network. Here, `R': reduction ratio of SRA; `H': head number of SRA; and  `E': expansion ratio of FFN}	
	\label{fig:fig_app1}
	\vspace{-0.1cm}
\end{figure}

Figure~\ref{fig:fig_app1} shows the detailed settings of the MiT network. 
The MiT encoder follows a progressive shrinking pyramid Transformer, as done in ~\cite{PVT}. It consists explicitly of four stages, each of which is composed of a SPE module and several stacked Transformers. 
In each stage, the SPE module down-samples the input features and generates the dimension-specific embedded sequence. 
Notably, we append an extra learnable SSL token~\cite{DINO,MoCov3} to the patch embedded sequence. 
The SSL token is similar to the [CLS] token in ViT, which is able to aggregate information from the whole patch embedding tokens via the self-attention.
The resultant sequences, combined with the learnable positional embedding, are inputted into the following Transformers for the long-term dependency modeling. 
Each Transformer layer includes a self-attention module and a feed-forward network (FFN) with two hidden layers. 
To reduce the computational cost and enable MiT to process high-resolution images, we follow the spatial-reduction attention (SRA) layer to reduce the spatial complexity~\cite{PVT}.
MiT has a symmetric decoder structure that consists of three stages. In each stage, the input feature map is first up-sampled by the SPE module, and then refined by the stacked Transformer layers.
Besides, we also add skip connections between the encoder and decoder to keep more low-level but high-resolution information. 
We devise two MiT by changing the number of Transformer layers, namely MiT-7 and MiT-22. Noticed that default MiT-22 is used in the main submission unless otherwise specified.

\begin{figure}[t!]
	\begin{center}
		{\includegraphics[width=0.6\linewidth]{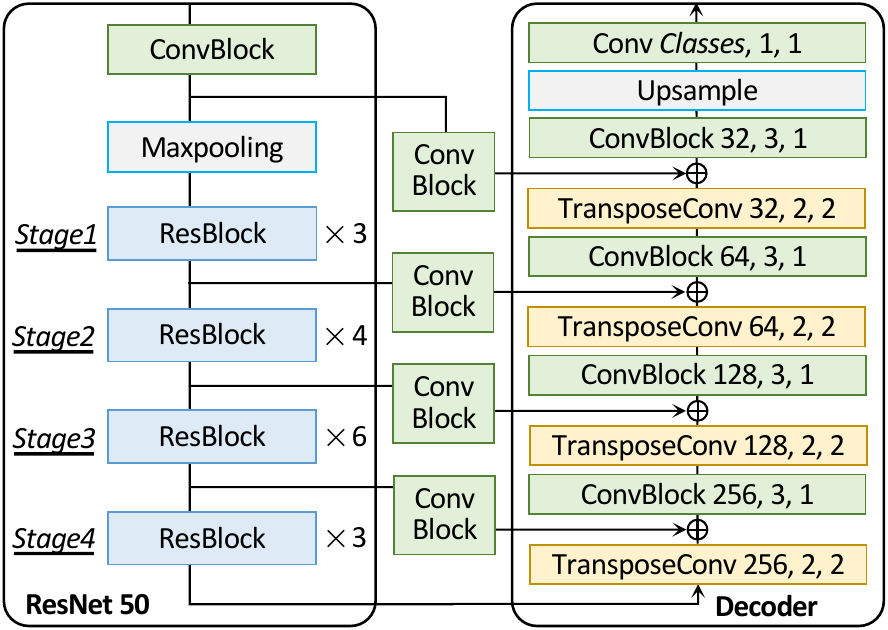}}
	\end{center}
	\vspace{-0.5cm}
	\caption{Detailed architecture of ResUnet: A 2D/3D ResNet-50~\cite{IN_pretrain} encoder, a decoder, and four skip connections between encoder and decoder. Green `ConvBlock': 2D Conv-Batch Normalization(BN)-ReLU or 3D Conv-IN-LeakyReLU; Yellow `TransConv': 2D/3D transposed convolutional layer. Note that the numbers in each block $/$ layer indicate the number of filters, kernel size, and stride, respectively.}	
	\label{fig:fig_app2}
	\vspace{-0.1cm}
\end{figure}

Figure~\ref{fig:fig_app2} shows the architecture of CNN-based ResUnet, used by the compared PCRL~\cite{PCRL}. It consists of a 2D/3D ResNet-50~\cite{IN_pretrain} encoder, a decoder, and four skip connections between encoder and decoder.
The decoder contains five up-sampling modules. Each of the first four modules has a transposed convolutional (TransConv) layer followed by a convolution block (ConvBlock) and a pixel-wise summation with the corresponding feature maps from the encoder and the TransConv layer. The last module comprises an Up-sampling layer followed by a $1\times1$ Conv layer that maps each 32-channel feature map to the desired number of classes.

\begin{figure}[t!]
	\begin{center}
		{\includegraphics[width=0.75\linewidth]{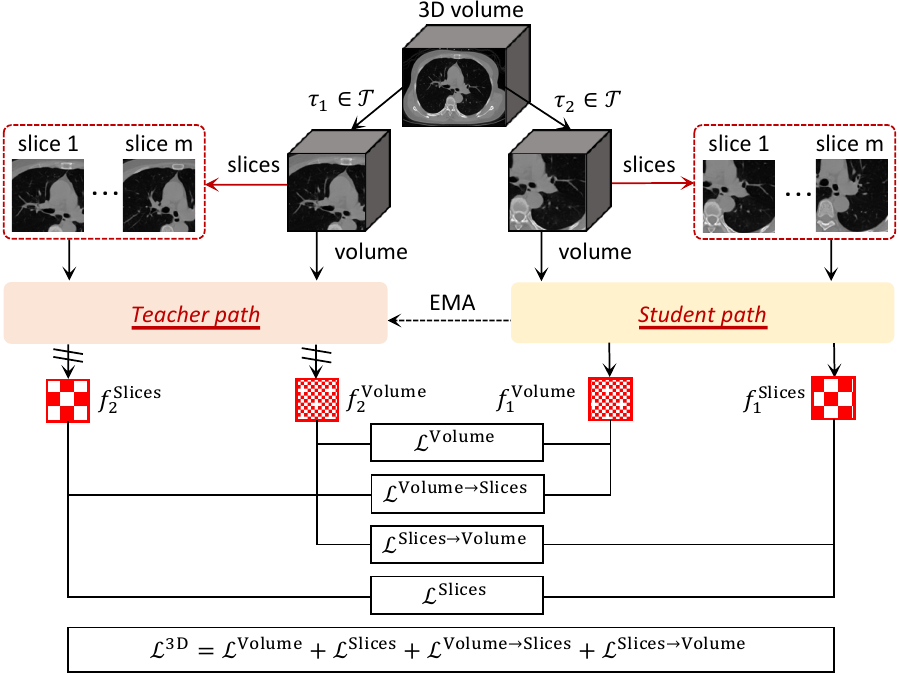}}
	\end{center}
	\vspace{-0.5cm}
	\caption{Intuitive explanation of volume-slice consistency mechanism.}	
	\label{fig:fig_app3}
	\vspace{-0.1cm}
\end{figure}

\section{Volume-slice consistency mechanism}
\label{Sec.Details_consist}

Figure~\ref{fig:fig_app3} gives an intuitive explanation of the proposed volume-slice consistency mechanism. 
Given a 3D volumetric image, we first create two augmented views via data augmentation, each of which has $m$ 2D slices. We then compute the volumetric or slice representations of dual paths, \ie $\bm{f}^{\mathrm{Volume}}_1$, $\bm{f}^{\mathrm{Volume}}_2$, $\bm{f}^{\mathrm{Slices}}_1$, and $\bm{f}^{\mathrm{Slices}}_2$. Here the slice representations $\bm{f}^{\mathrm{Slices}}_1$ and $\bm{f}^{\mathrm{Slices}}_2$ are generated by averaging the outputs of $m$ slices. 
The loss function is composed of four items, including $\mathcal{L}^{\mathrm{Volume}}$, $\mathcal{L}^{\mathrm{Slices}}$, $\mathcal{L}^{\mathrm{Volume \to Slices}}$, and $\mathcal{L}^{\mathrm{Slices \to Volume}}$. 
The first two items aim to achieve the consistency at the level of global volume and local slices, respectively. 
Besides, the consistency across both levels should also be satisfied, which is achieved by the latter two items. 
By jointly using these four loss items, our model is able to capture richer representations from 3D medical images.

\end{document}